\documentclass[11pt]{article}

\usepackage[]{acl}

\usepackage{times}
\usepackage{latexsym}
\usepackage[T1]{fontenc}
\usepackage[utf8]{inputenc}
\usepackage{microtype}
\usepackage{inconsolata}

\usepackage{graphicx}
\usepackage{array}
\usepackage{tabularx}
\usepackage{booktabs}
\usepackage{multirow}
\usepackage{makecell}       %
\usepackage{arydshln}       %

\usepackage{xcolor}         %
\usepackage{pifont}         %
\usepackage[most]{tcolorbox}%
\usepackage[normalem]{ulem} %

\usepackage{enumitem}       %
\usepackage{seqsplit}       %

\newcolumntype{M}{>{\raggedright\arraybackslash}p{10em}}        %
\newcolumntype{Y}{>{\raggedright\arraybackslash}X}              %
\newcolumntype{H}[1]{>{\def\colhue{#1}\centering\arraybackslash}X} %

\definecolor{cSupportedFg}{HTML}{16537E}
\definecolor{cRefutedFg}{HTML}{9C2B20}
\definecolor{hueA}{HTML}{2A9D8F} %
\definecolor{hueC}{HTML}{3B7DD8} %
\definecolor{hueE}{HTML}{E08A1E} %

\newcommand{\yessym}{\ding{51}}
\newcommand{\ptsym}{$\triangle$}
\newcommand{\nosym}{\textcolor{black!35}{--}}

\newcommand{\rowstrut}{\rule[-\arraystretch\dp\strutbox]{0pt}%
  {\dimexpr\arraystretch\ht\strutbox+\arraystretch\dp\strutbox\relax}}
\newcommand{\fillcell}[2]{\makebox[\linewidth][c]{\setlength{\fboxsep}{0pt}%
  \colorbox{#1}{\makebox[\dimexpr\linewidth+2\tabcolsep\relax][c]{\rowstrut#2}}}}
\newcommand{\cellY}{\fillcell{\colhue!45}{\yessym}}
\newcommand{\cellP}{\fillcell{\colhue!16}{\ptsym}}
\newcommand{\cellN}{\nosym}

\newcommand{\groupgap}{\hspace{\dimexpr2\tabcolsep+3pt\relax}}

\newcommand{\bench}[2]{\hyperlink{cite.#1}{#2}~\citeyearpar{#1}}

\newcommand{\cmark}{\ding{51}}
\newcommand{\xmark}{\textcolor{gray!60}{\ding{55}}}

\newcommand{\tagS}[2]{%
  \tcbox[on line, box align=base,
    arc=2pt, outer arc=2pt,
    colback=#1!12, colframe=#1!12,
    boxrule=0pt, boxsep=0pt,
    left=2.5pt, right=2.5pt, top=1.5pt, bottom=1.5pt]%
    {\textcolor{#1!90!black}{\footnotesize\bfseries #2}}}
\newcommand{\badgeSupported}{\tagS{cSupportedFg}{\ding{51}\,Supported}\,}
\newcommand{\badgeRefuted}{\tagS{cRefutedFg}{\ding{55}\,Refuted}\,}
\newcommand{\badgeUnv}{\tagS{black}{?\,NotEnoughInfo}\,}

\newcommand{\impB}[2]{%
  \tcbox[on line, box align=base,
    arc=2pt, outer arc=2pt,
    colback=white, colframe=black!#1,
    boxrule=0.5pt, boxsep=0pt,
    left=2.5pt, right=2.5pt, top=1.5pt, bottom=1.5pt]%
    {\textcolor{black!#1}{\footnotesize\bfseries #2}}}
\newcommand{\impMain}{\impB{75}{Main}}
\newcommand{\impSub}{\impB{55}{Sub}}
\newcommand{\impIrr}{\impB{35}{Irrelevant}}

\newcommand{\claimline}[2]{%
  {\setlength{\ULdepth}{0.55ex}\renewcommand{\ULthickness}{1.1pt}%
   \normalcolor
   \begingroup\color{#1}%
   \uline{#2}%
   \endgroup}}
\newcommand{\claimX}[3]{\claimline{#1}{#2}\nobreak\kern1pt#3}
\newcommand{\claimT}[1]{\claimX{cSupportedFg}{#1}{\badgeSupported}}
\newcommand{\claimF}[1]{\claimX{cRefutedFg}{#1}{\badgeRefuted}}

\newcommand{\rowlab}[1]{\textbf{\small #1}}
\newcommand{\claimlab}[1]{\hspace{0.8em}\textbf{\small #1}}
\newcommand{\midelide}{{\color{black!45}\,$\cdots$\,}}

\newcommand{\evidence}[3]{%
  \par\vspace{2pt}%
  \noindent\hspace*{1em}%
  \parbox{\dimexpr\linewidth-1em\relax}{%
    \footnotesize\color{black!55}%
    \textbf{Source ID:}\,\texttt{#1},\;\textbf{Time:}\,\texttt{#2}\par
    \textbf{Evidence span:}\,#3\par
  }%
  \par\vspace{2pt}%
}

\title{Rethinking the Evaluation of LLM Fact-Checkers}

\author{%
  \textbf{Ayana Niwa} \\
  MBZUAI \\
  {\fontsize{9.5}{11}\texttt{ayana.niwa@mbzuai.ac.ae}}
\And
  \textbf{Miyu Yamada} \\
  Science Tokyo \\
  {\fontsize{9.5}{11}\texttt{yamada.m.ba@m.titech.ac.jp}}
\And
  \textbf{Kouta Nakayama} \\
  NII-LLMC \\
  {\fontsize{9.5}{11}\texttt{nakayama@nii.ac.jp}}
\AND
  \textbf{Masahiro Kaneko} \\
  MBZUAI \\
  {\fontsize{9.5}{11}\texttt{masahiro.kaneko@mbzuai.ac.ae}}
\And
  \textbf{Hisami Suzuki} \\
  NII-LLMC \\
  {\fontsize{9.5}{11}\texttt{hisamis@nii.ac.jp}}
\And
  \textbf{Satoshi Sekine} \\
  NII-LLMC \\
  {\fontsize{9.5}{11}\texttt{sekine@nii.ac.jp}}
}

\begin{document}
\maketitle

\begin{abstract}
Large language models (LLMs) are widely used to answer factual questions, but their answers can contain errors and must be fact-checked claim by claim.
Yet the resulting score depends on three design decisions: which questions the claims answer, under what conditions each claim is judged, and which sources count as evidence.
No existing benchmark specifies all three.
We introduce \textbf{JSFactCheckBench}\footnote{\url{https://huggingface.co/datasets/llm-jp/JSFactCheckBench}}, a Japanese benchmark that specifies all three, with 1,913 claims from LLM answers to questions about misleading information that circulated on Japanese social media, each annotated for factuality at a reference time and for importance to the question, using evidence from preselected Web sources.
Across six LLMs, all verify claims directly answering the question less accurately than supplementary claims, a gap that aggregate scores obscure.
Scores also drop when models must locate the relevant content in longer or multiple sources and, for all four open models, when the surrounding question and response are added.
These weaknesses are visible only because the benchmark records the conditions behind each label. A benchmark that leaves them implicit cannot reveal them.
\end{abstract}

\begin{table*}[t]
\centering
\small
\setlength{\tabcolsep}{3pt}
\renewcommand{\arraystretch}{1.15}
\begin{tabularx}{\textwidth}{
l
l
@{\groupgap} H{hueA} H{hueA}
@{\groupgap} H{hueC} H{hueC}
@{\groupgap} H{hueE} H{hueE}
}
\toprule
& & \multicolumn{2}{c}{\textbf{(1) Claim collection}}
& \multicolumn{2}{c}{\textbf{(2) Judgment conditions}}
& \multicolumn{2}{c}{\textbf{(3) Evidence sources}} \\
\cmidrule(lr){3-4}
\cmidrule(lr){5-6}
\cmidrule(lr){7-8}
Benchmark
& Lang.
& \makecell{LLM\\responses}
& \makecell{Real-world\\origin}
& \makecell{Reference\\time}
& \makecell{Human\\importance}
& \makecell{Open\\web}
& \makecell{Source\\control} \\
\midrule
\bench{thorne-etal-2018-fever}{FEVER}
& En & \cellN & \cellN & \cellN & \cellN & \cellN & \cellY \\
\bench{lin-etal-2022-truthfulqa}{TruthfulQA}
& En & \cellP & \cellN & \cellN & \cellN & \cellY & \cellN \\
\bench{hu-etal-2022-chef}{CHEF}
& Zh & \cellN & \cellY & \cellP & \cellN & \cellY & \cellP \\
\bench{singhal-etal-2022-factdrill}{FactDrill}
& Multi & \cellN & \cellY & \cellP & \cellN & \cellY & \cellN \\
\bench{min-etal-2023-factscore}{FActScore}
& En & \cellY & \cellN & \cellN & \cellN & \cellN & \cellY \\
\bench{schlichtkrull2023averitec}{AVeriTeC}
& En & \cellN & \cellY & \cellP & \cellN & \cellY & \cellN \\
\bench{chen2023felm}{FELM}
& En & \cellY & \cellP & \cellN & \cellN & \cellY & \cellN \\
\bench{10825770}{JSocialFact}
& Ja & \cellN & \cellY & \cellN & \cellN & \cellN & \cellN \\
\bench{wang-etal-2024-factcheck}{Factcheck-Bench}
& En & \cellY & \cellP & \cellN & \cellP & \cellY & \cellN \\
\bench{vazquez-etal-2025-semeval}{Mu-SHROOM}
& Multi & \cellY & \cellN & \cellN & \cellN & \cellN & \cellY \\
\bench{wanner2025claimsequalclaimsequal}{VITALERRORS}
& En & \cellP & \cellP & \cellN & \cellP & \cellN & \cellY \\
\bench{yao2025factcheckingaigeneratednewsreports}{Yao et al.}
& En & \cellY & \cellN & \cellP & \cellN & \cellY & \cellN \\
\bench{rothermel-etal-2026-veritas}{VeriTaS}
& Multi & \cellN & \cellY & \cellP & \cellN & \cellN & \cellN \\
\bench{sasaki2026jadafc}{JAD-AFC}
& Ja & \cellN & \cellY & \cellP & \cellN & \cellY & \cellN \\
\midrule
\textbf{JSFactCheckBench}
& Ja & \cellY & \cellY & \cellY & \cellY & \cellY & \cellY \\
\bottomrule
\end{tabularx}
\caption{
Comparison of construction. \yessym: satisfied; \ptsym: partially satisfied; \nosym: not satisfied.
\textit{LLM responses}: factuality or hallucination annotations of LLM-generated responses at claim, segment, or span level (\ptsym: answer-level labels or perturbation conditions only).
\textit{Real-world origin}: provenance of the source questions or claims (\ptsym: mixed provenance). VeriTaS and JAD-AFC include truth-altering claim variants.
\textit{Reference time}: a recorded reference time explicitly anchoring gold factuality judgments (\ptsym: temporal metadata or constraints only).
\textit{Human importance}: manually annotated claim importance that is incorporated into the benchmark's evaluation (\ptsym: annotated but not used in evaluation, or automatically assigned).
\textit{Open web}: Web-derived evidence or reference sources not restricted to a fixed reference corpus.
\textit{Source control}: predefined restrictions on eligible evidence sources (\yessym: fixed reference sources or an allowlist;
\ptsym: blocklist-based restrictions).
AVeriTeC excludes only the originating fact-checking article and merely flags unreliable sites.
}
\label{tab:comparison}
\end{table*}

\begin{figure*}[t]
    \centering
    \includegraphics[width=\linewidth]{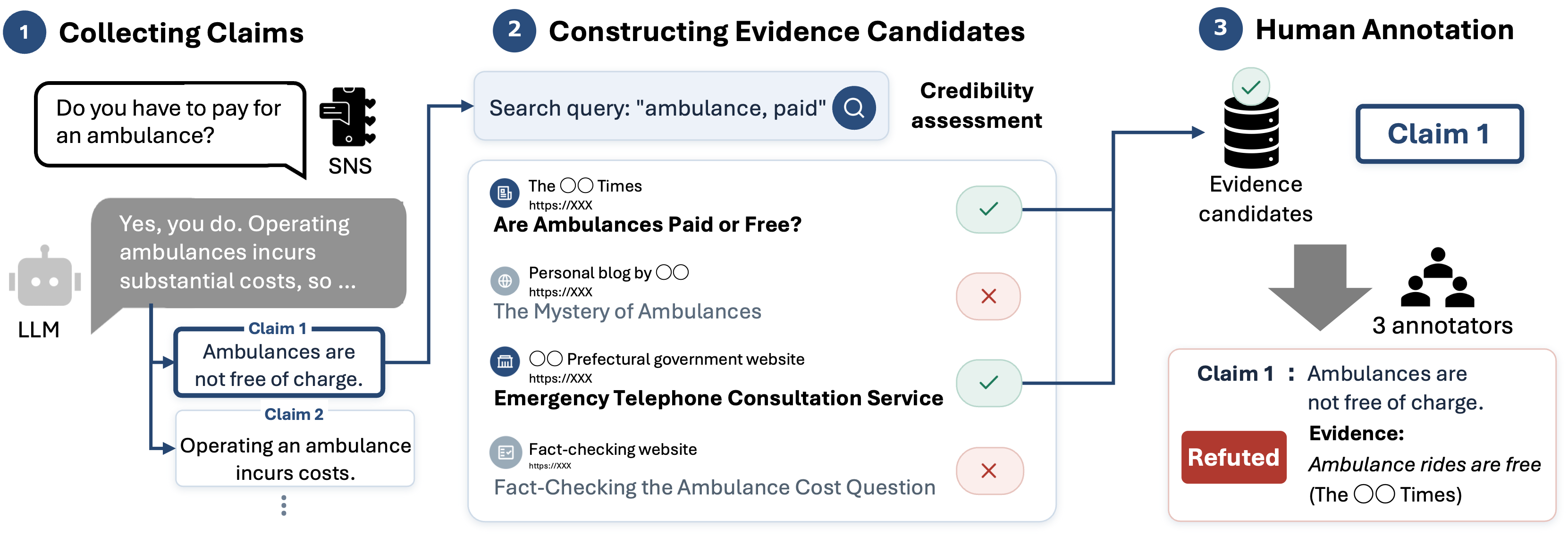}
    \caption{Overview of the dataset construction pipeline.}
    \label{fig:pipeline}
\end{figure*}

\section{Introduction}\label{sec:intro}

Large language models (LLMs) are increasingly used as knowledge sources to answer factual questions~\cite{petroni-etal-2019-language,mallen-etal-2023-trust}.
However, flawed training data and missing or outdated knowledge can lead them to produce false or misleading information~\cite{lin-etal-2022-truthfulqa,kadavath2022languagemodelsmostlyknow}.
Automated fact-checking of LLM responses is therefore important.
Because a response may contain several claims, fact-checking means verifying each claim it contains, and evaluating such systems requires reliable benchmarks.

Three design decisions determine what a benchmark measures, and if any is left uncontrolled, the labels depend on it implicitly.
\textbf{(1) Which claims to verify.}
A benchmark's questions should come from the real world, since questions constructed for evaluation may not reflect what arises in practice.
Its scores then describe performance for the population those questions come from.
\textbf{(2) Under what conditions to judge them.}
The truth of many claims changes, so labels should be tied to a reference time.
Without one, a label holds as of whenever the annotator happened to search, a date the verifier is never told.
An error in a claim directly answering the question also matters more than one in a peripheral detail, so labels should be tied to the claim's importance.
Without it, a wrong answer to the question weighs the same as a wrong aside.
\textbf{(3) Which sources to accept as evidence.}
Web pages vary in reliability and may fact-check the claim, leaking the verdict, so sources should be controlled.
Otherwise the evidence is whatever page the annotator found, reliable or not.
No existing benchmark controls all three (\autoref{tab:comparison}).

We introduce \textbf{JSFactCheckBench}, a Japanese fact-checking benchmark designed to make all three decisions explicit (\autoref{fig:pipeline}).
For (1), we draw on JSocialFact~\cite{10825770}, whose questions derive from misleading information that actually circulated on Japanese social media, and extract 1,913 claims from the answers of eleven LLMs to them.
For (2), each claim is annotated for factuality at a specified reference time and for importance to the question.
For (3), factuality is grounded in evidence from preselected Web sources, and three professional annotators independently judge each claim.

Decision (1) is settled once the data are collected.
The other two remain open at evaluation time, and the benchmark shows what each changes.
For (2), only a quarter of claims directly answer the question, and all six LLMs we evaluate verify those claims less accurately than the supplementary ones, a gap that aggregate scores hide where it matters most.
Answers to 46\% of the questions can also change over time, and stating the reference time shifts the scores of some models but not others.
For (3), because every candidate page is recorded, we can ask models to select the source rather than read a given passage, and the spread between models triples when they must.
None of this is visible from a single score under a single condition, and none of it can be measured on a benchmark that leaves these decisions uncontrolled.

\section{JSFactCheckBench}\label{sec:dataset}

\subsection{Dataset Design}\label{sec:dataset_design}

JSFactCheckBench evaluates LLM answers to factual questions.
Because a response may contain several claims differing in factuality and importance to the question, the unit of evaluation is an individual claim rather than the response as a whole.

The benchmark follows three design principles corresponding to the decisions introduced in \autoref{sec:intro}: (1) Claims come from LLM answers to questions of real-world origin. (2) Factuality is judged at a stated reference time, and each claim is labeled for its importance to the question. (3) Web evidence is collected under a controlled procedure.
\autoref{fig:pipeline} gives an overview of the construction process that implements these principles, and \autoref{tab:example} shows examples from our dataset.

\paragraph{(1) LLM claims to real questions.}
To evaluate the errors that LLMs actually make, a benchmark should draw its claims from LLM responses to questions of real-world origin, rather than from claims constructed for evaluation.
The benchmark is built for users in Japan, so we take questions about misleading information that actually circulated on Japanese social media and extract claims from LLM answers to them.

\begin{table*}[t]
  \centering
  \small
  \setlength{\tabcolsep}{0pt}
  \renewcommand{\arraystretch}{1.1}
  \begin{tabularx}{\textwidth}{@{}l@{\hspace{1.2em}}X@{}}
    \toprule
    \rowlab{Question 1} & Is it free to call an ambulance in Japan?\\
    \midrule
    \rowlab{Response} & Calling an ambulance in Japan normally incurs a fee. However, \midelide{} to request an ambulance, call 119. \midelide\\\hdashline
    \addlinespace[3pt]
    \claimlab{Claim} & Calling an ambulance in Japan normally \claimF{incurs a fee.}\,\impMain%
      \evidence{d61417498a}{2026-02-15}%
               {In Japan, ambulance users currently bear no charge and can use the service free of cost.\midelide{}}\\
    \midrule
    \rowlab{Question 2} & Will the ¥1.1 million gift tax exemption be abolished starting in 2024?\\\midrule 
    \rowlab{Response} & No decision to abolish the annual gift tax exclusion of ¥1.1 million in 2024 has been announced. The tax-free threshold for gift tax is set under Japanese tax legislation. \midelide{} \\\hdashline
    \addlinespace[3pt]
    \claimlab{Claim} & \strut\claimT{No decision to abolish the ¥1.1 million gift tax exemption in 2024 has been announced.}\,\impMain%
      \evidence{b3655be520}{2026-02-15}%
               {\midelide{} it was decided that the annual gift tax system would remain in place.\midelide{}}\\
    \bottomrule
  \end{tabularx}
  \caption{Example instances from JSFactCheckBench. The Japanese text is translated into English.}
  \label{tab:example}
\end{table*}

\paragraph{(2) Labels conditioned on time and the question.}
To make factuality judgments well-defined and distinguish errors by their consequence for the answer, we condition each claim on  two aspects.

\underline{\textbf{(2)-A. Factuality at a specified time.}}
To assign a well-defined label to claims whose truth may change, we define factuality with respect to a specified reference time.
For example, for the claim ``\textit{Alice is the CEO of Company X}'' and the reference time ``\textit{January 2025},'' the label depends on whether it was true at that time, not at the time of evaluation.
We therefore provide the reference time to the verifier and label each claim \badgeSupported if the evidence supports it at that time, \badgeRefuted if it contradicts it, and \badgeUnv if it does neither.

\underline{\textbf{(2)-B. Importance to the question.}}
To distinguish errors in the direct answer itself from errors in supplementary content, we label each claim by its role in answering the question.
For the question ``\textit{Is it free to call an ambulance in Japan?}'', a claim is \impMain{} if it directly answers the question or states the core or part of the answer (``\textit{Calling an ambulance is generally free of charge.}''), \impSub{} if it provides related information without directly answering the question (``\textit{Medical treatment at the receiving hospital may incur a fee.}''), and \impIrr{} if it has no direct relation to the question or answer (``\textit{Fire engines in Japan are red.}'').

\paragraph{(3) Controlled Web evidence.}\label{sec:evidence_design}
To verify open-domain claims while keeping gold labels grounded in accountable evidence, the benchmark must combine broad Web coverage with control over which sources can determine factuality.
We therefore restrict evidence candidates to a predefined domain allowlist of accountable sources and exclude two fact-checking-site domains to reduce verdict leakage.
For each claim, all annotators inspect the same pre-collected candidate set and record the URL and passage used for their judgment.

\subsection{Dataset Construction}

We build the dataset in three steps: collecting claims from LLM responses, constructing web evidence candidates, and human annotation.

\paragraph{Step 1: Collecting claims.}\label{sec:statement}\leavevmode\par

We randomly sampled 85 of the 385 questions in JSocialFact~\cite{10825770}, a benchmark built from misleading information that circulated on Japanese social media.
For each question, we generated two responses, 170 in total, using two distinct models sampled at random from a pool of eleven (Appendix~\ref{app:model_list}).
This enables us to cover a broad range of factual errors.

Following \citet{liu-etal-2025-verifact}, we decompose each response into atomic claims, that is, statements that contain a single proposition and can be verified independently of the rest of the response.
We apply their prompt with Meta-Llama-3-8B-Instruct.
The procedure (1) splits each sentence into atomic claims; (2) resolves ambiguous references such as pronouns and incomplete entity names; (3) restores omitted temporal, causal, and comparative relations; (4) adds contextual information such as presuppositions; and (5) checks and revises each claim for concrete, objective verifiability.
The LLM performs all five steps automatically.

\paragraph{Step 2: Constructing evidence candidates.}\label{sec:evidence}
\leavevmode\par
\textbf{Selecting trusted domains.}
Since evidence determines gold labels, which sources may determine them must be fixed in advance rather than left to each annotator.
We therefore admit only sources that an institution outside the benchmark has vetted.
This excludes anonymous and unvetted material and makes the choice reproducible across annotators.
Because this cannot be judged page by page, we apply it at the domain level and compile an allowlist, fixed before retrieval, from four groups, each defined by a different vetting institution:

\begin{itemize}[leftmargin=1.2em, topsep=3pt, itemsep=2pt, parsep=0pt]
     \item \textbf{Core.} Domains of government (\texttt{go.jp}), local government (\texttt{lg.jp}), academic institutions (\texttt{ac.jp}), companies (\texttt{co.jp}), and incorporated organizations (\texttt{or.jp}), which the registry issues only to verified organizations.
    \item \textbf{Media.} Domains of 73 organizations listed by the Japan Newspaper Publishers \& Editors Association, including newspapers, broadcasters, advertising companies, and publishers,\footnote{\url{https://www.pressnet.or.jp/medialink/}} which are bound by its editorial standards.
    \item \textbf{Companies.} Domains of 1,029 companies listed on the Tokyo Stock Exchange, collected from Wikipedia. These add official corporate sites outside \texttt{co.jp}, whose statements about their own products, services, and announcements are subject to listing rules.
    \item \textbf{Wikipedia references.} 322 domains that appeared at least twice as references on Japanese Wikipedia pages about politics, incidents, disasters, and elections. These are sources that Wikipedia's sourcing policy, enforced by its editors, has repeatedly admitted, and they add what the other groups miss.
\end{itemize}

In compiling these groups, we excluded two fact-checking sites, whose articles may state an existing verdict for a target claim.
We also excluded domains listed on Wikipedia pages for blogs and microblogging services, as well as \texttt{note.com}, judging these sources unsuitable as benchmark evidence because author identity and editorial responsibility are not consistently established.
The resulting allowlist contains 1,397 domains and is included in the released dataset.
A domain belonging to multiple groups is counted under the highest-priority group, in the order listed above; the same rule applies to the per-category counts in Table~\ref{tab:evidence_domain}.
The allowlist is specific to Japan because the benchmark targets users there. The procedure itself is not.
For another target, one compiles an allowlist of accountable sources for that region similarly.

\textbf{Searching evidence candidates.}
For each context-completed claim from Section~\ref{sec:statement}, an LLM generated a keyword query meeting three requirements: (1) it should retrieve information to verify concrete facts in the claim, including dates, quantities, named entities, and events; (2) it should contain expressions likely to occur in government releases, news articles, or official statistics; and (3) it should include enough context to distinguish the target from similarly named entities or events.
The prompt appears in Appendix~\ref{app:query}.
We submitted each query to the Custom Search JSON API\footnote{\url{https://developers.google.com/custom-search/v1/overview}}, set the search region to Japan, and retrieved the top 30 results.
We retained only results from allowlisted domains, excluded \texttt{.pdf} URLs, and selected up to 10 top-ranked evidence candidates per claim.

\paragraph{Step 3: Human Annotation.}\label{sec:annotation}\leavevmode\par

Annotators assign the factuality and importance labels defined in (2)-A and (2)-B of Section~\ref{sec:dataset_design}. 
Claims that are not truth-evaluable, such as metalinguistic statements or subjective opinions, receive the auxiliary label \textit{Unevaluable}; malformed or uninterpretable claims receive \textit{Unclear}.
Full guidelines are in Appendix~\ref{app:guideline}.

Three annotators used Label Studio\footnote{\url{https://labelstud.io/}} to read the question and full response, inspect the evidence candidates, and label each target claim.
For Refuted claims, annotators additionally marked the erroneous span; for all other labels, the annotation covered the full claim.
Annotators also recorded the evidence URL and the passage used for the factuality judgment.
We then consolidated the three annotations into one gold factuality label and one gold importance label per claim.
For factuality, a Supported or Refuted judgment takes precedence over NotEnoughInfo because it is accompanied by evidence supporting or contradicting the claim.
When annotations contain both NotEnoughInfo and either Supported or Refuted, we retain the latter; when Supported and Refuted conflict, we use majority vote.
Unclear and Unevaluable are retained only when all three annotators assign them; otherwise, they are ignored.
Importance labels are consolidated by majority vote.
This automatic consolidation procedure resolved 98\% of factuality labels and 97\% of importance labels.
For the remaining claims, two additional annotators discussed the case and assigned the final label.
Because claims in LLM answers have no existing verdicts, each was verified from scratch, at about USD~7.50 per claim; Appendix~\ref{app:annotation_cost} reports the cost.

To validate the labels, two additional annotators independently relabeled 50 sampled claims, given the claim and the passage used in the original judgment; all three label sets agreed on 94\% of the claims for factuality and 96\% for importance.

\subsection{Dataset Statistics}\label{sec:stats}
JSFactCheckBench contains 1,913 claims from 170 responses to 85 questions, with 11.3 claims per response on average (range 1--24; Figure~\ref{fig:claims}).
These responses broadly cover the outputs of the eleven models, as shown in Appendix~\ref{app:model_list}.

\paragraph{Label distribution.}

\begin{table}[t]
\centering\small
\setlength{\tabcolsep}{5pt}
\resizebox{\linewidth}{!}{
\begin{tabular}{@{}lrrrr@{}}
\toprule
& \multicolumn{1}{c}{Main} & \multicolumn{1}{c}{Sub} & \multicolumn{1}{c}{Irrel.} & \multicolumn{1}{c}{All} \\
\midrule
Supported     & 250 & 625 & 54 & 929 (48.6\%) \\
Refuted       & 70 & 172 & 8 & 250 (13.1\%) \\
NotEnoughInfo & 140 & 345 & 74 & 559 (29.2\%) \\
Unevaluable   & 23 & 108 & 44 & 175 \phantom{0}(9.1\%) \\
\midrule
All & 483 & 1,250 & 180 & 1,913 \\
    & (25.2\%) & (65.3\%) & (9.4\%) & \\
\bottomrule
\end{tabular}
}
\caption{Label distribution of factuality and importance. Percentages are relative to all 1,913 claims; the 20 claims labeled Unclear are excluded from the dataset.}
\label{tab:crosstab}
\end{table}

Table~\ref{tab:crosstab} shows the label distribution of factuality and importance. 
Among claims that received either Supported or Refuted, 21\% are Refuted, indicating that roughly one in five verifiable claims in these responses is incorrect. 
On the importance axis, Sub claims account for 65.3\% of the data and Main claims for 25.2\%. Most of an LLM answer therefore consists of supplementary material rather than statements that directly answer the question, which is precisely the situation in which an unweighted claim-level error rate becomes difficult to interpret.

\paragraph{Question attributes.}

We classified the 85 questions along four dimensions: topic domain, geographic scope, question stance, and temporal stability.
Tables~\ref{tab:taxonomy_axes1}--\ref{tab:taxonomy_axes5} define the taxonomy.

\begin{table}[t]
\centering\footnotesize
\setlength{\tabcolsep}{0pt}
\begin{tabular*}{\columnwidth}{@{\extracolsep{\fill}}lrlr@{}}
\toprule
Topic domain & N & Topic domain & N \\
\midrule
Medicine and health     & 24 & Science and technology  & 4 \\
Nuclear energy          & 12 & Justice and law         & 4 \\
Economy and industry    & 12 & Society and gender      & 4 \\
Culture and history     &  8 & Politics                & 3 \\
Government services     &  7 & International affairs   & 2 \\
Media and broadcasting  &  5 &                         &   \\
\bottomrule
\end{tabular*}
\caption{Distribution of question domains.}
\label{tab:domain}
\end{table}

\begin{table}[t]
\centering\small
\setlength{\tabcolsep}{5pt}
\begin{tabular}{@{}lrr@{}}
\toprule
Geographic scope & \multicolumn{1}{c}{N} & \multicolumn{1}{c}{\%} \\
\midrule
Japan                 & 40 & 47.1 \\
Universal/non-geographic & 24 & 28.2 \\
Foreign country       & 14 & 16.5 \\
International/global  & 7 & 8.2 \\
\bottomrule
\end{tabular}
\caption{Distribution of geographic scope.}
\label{tab:geo}
\end{table}

\textbf{Domain.}
The questions span 11 domains (Table~\ref{tab:domain}).
\textit{Medicine and health} is the largest at 28.2\%, followed by \textit{nuclear energy and radiation} and \textit{economy, industry, and business} (14.1\% each).
This concentration likely reflects the topics dominating misleading information on Japanese social media.
\textbf{Geography.}
Of the questions, 63.5\% concern a specific country, and 47.1\% overall concern Japan (Table~\ref{tab:geo}).
Questions about Japanese institutions and events may require domestic evidence such as government publications and news reports.
JSFactCheckBench restricts its questions by origin (Japanese-language claims), not by subject matter, so foreign and non-geographic topics are included.
\textbf{Stance.}
Most questions are posed neutrally (83.5\%; e.g., ``\textit{Did Azerbaijan originate from Turkey?}''), while 10.6\% attribute the proposition to another source (e.g., ``\textit{Is it true that the vegan meal in JAL business class was only one banana?}'') and 5.9\% presuppose it and invite agreement (e.g., ``\textit{Isn't it strange to say that ordinary water molecules and tritiated water molecules are the same?}'').
\textbf{Temporal stability.}
Temporally dynamic questions account for 46\%.
Their answers change with company policy (``\textit{Does Apple Vision Pro have an F1 app?}'') or regulatory action (``\textit{Has the Consumer Affairs Agency recognized the effectiveness of Cleverin?}''), unlike static scientific questions (``\textit{Does drinking baking soda cure cancer?}'').

\paragraph{Evidence distribution.}
Each claim has 9.42 evidence candidates on average, and only 2 claims (0.1\%) have no candidate from an allowlisted domain.
\badgeUnv labels therefore rarely arise from empty retrieval; in almost all cases annotators had candidates to inspect but could not establish the claim's truth or falsity from them.
Because the same page can appear as a candidate for multiple claims, the candidate sets comprise 10,011 unique pages, of which annotators selected 667 unique pages (6.7\%) as evidence for their judgments.
Government sources are selected disproportionately often: within the Core group, \texttt{go.jp} accounts for 30.4\% of candidates but 38.0\% of selected links, whereas \texttt{co.jp} supplies a comparable share of candidates (34.8\%) and a smaller share of selections (31.9\%).
The most frequently selected domain was the Ministry of Health, Labour and Welfare (\texttt{mhlw.go.jp}; 23 pages), followed by the Asahi Shimbun (\texttt{asahi.com}; 20) and the Ministry of Foreign Affairs (\texttt{mofa.go.jp}; 15).
Table~\ref{tab:evidence_domain} describes the distribution of domain categories.

\subsection{Comparison with Existing Benchmarks}
\label{sec:related}

Table~\ref{tab:comparison} compares dataset construction along the three design decisions.
JSFactCheckBench combines all six properties defined in the caption.

\paragraph{Claim collection.}
FEVER and TruthfulQA use annotator-written claims or reference answers, while FactDrill, CHEF, AVeriTeC, VeriTaS, and JAD-AFC draw claims from fact-checking articles rather than LLM answers. 
Among LLM-answer datasets, FActScore focuses on biographies, Mu-SHROOM constructs questions from Wikipedia, and FELM, Factcheck-Bench, and VITALERRORS mix prompt sources. 
\citet{yao2025factcheckingaigeneratednewsreports} extract claims from news reports generated from deliberately inconsistent scenarios rather than real-world questions.
The other Japanese resources do not release claim-level labels on LLM answers; our claims come from LLM answers to JSocialFact questions derived from misleading information in Japan.

\paragraph{Judgment conditions.}
None of the compared datasets records a reference time that anchors gold labels.
AVeriTeC's evidence cutoff and VeriTaS's dated, quarterly claims use the date a claim was made to keep later information out, CHEF writes missing times into the claim, and \citet{yao2025factcheckingaigeneratednewsreports} give the verifier a fixed date without recording one for the human labels.
We instead fix when the label must hold: the evidence-collection date, unless the claim specifies a time, with only evidence available by then, and we give that date to the verifier.
Factcheck-Bench annotates importance manually but reports it only as dataset statistics, and VITALERRORS assigns it automatically, to edited text.
We manually annotate question-relative importance in unedited LLM answers and use it in evaluation.

\paragraph{Evidence.}
FEVER, FActScore, and Mu-SHROOM use Wikipedia, while VITALERRORS uses reference documents from its source datasets, limiting evidence coverage to those sources.
Open-Web datasets generally lack a predefined domain allowlist; CHEF instead blocklists misinformation and fact-checking sites.
A blocklist still admits any unvetted site it does not list, so we use an allowlist, exclude two fact-checking-site domains to reduce verdict leakage~\cite{glockner-etal-2022-missing,chrysidis2024credible}, and provide all annotators with the same stored evidence candidates.

\section{Experiments}\label{sec:experiment}

\begin{table*}[t]
\centering
\small
\setlength{\tabcolsep}{5pt}
\begin{tabular}{ccclcccccc}
\toprule
\multicolumn{3}{c}{Input given to the model} & & \multicolumn{6}{c}{Macro-F1} \\
\cmidrule(lr){1-3}\cmidrule(lr){5-10}
Claim & Time & Q\,\&\,R & Evidence & Llama-3.1-Sw. & Qwen3-Sw. & LLM-jp-4 & Mistral-Nemo-Ja & GPT-5.1 & Sonnet~5 \\
\midrule
\multicolumn{10}{l}{\textit{(a) Closed book: no evidence supplied}}\\
\midrule
\cmark & \xmark & \xmark & None      & 0.444 & 0.463 & 0.485 & 0.490 & 0.549 & \textbf{0.600} \\
\cmark & \cmark & \xmark & None      & 0.418 & 0.437 & 0.474 & 0.489 & 0.557 & \textbf{0.593} \\
\cmark & \cmark & \cmark & None      & 0.265 & 0.327 & 0.307 & 0.256 & 0.544 & \textbf{0.586} \\
\midrule
\multicolumn{10}{l}{\textit{(b) Source given: the cited page is supplied}}\\
\midrule
\cmark & \xmark & \xmark & Ref. & 0.691 & 0.695 & 0.752 & 0.724 & \textbf{0.770} & 0.753 \\
\cmark & \cmark & \xmark & Ref. & 0.684 & 0.666 & 0.741 & 0.745 & \textbf{0.776} & 0.761 \\
\cmark & \cmark & \cmark & Ref. & 0.350 & 0.352 & 0.370 & 0.334 & 0.714 & \textbf{0.750} \\
\addlinespace[3pt]
\cmark & \xmark & \xmark & Full      & 0.569 & 0.670 & 0.670 & 0.629 & 0.701 & \textbf{0.718} \\
\cmark & \cmark & \xmark & Full      & 0.582 & 0.654 & 0.675 & 0.664 & 0.716 & \textbf{0.726} \\
\cmark & \cmark & \cmark & Full      & 0.344 & 0.392 & 0.422 & 0.340 & 0.707 & \textbf{0.714} \\
\midrule
\multicolumn{10}{l}{\textit{(c) Search-like: the source must be selected}}\\
\midrule
\cmark & \xmark & \xmark & Pages     & 0.458 & 0.622 & 0.624 & 0.581 & 0.700 & \textbf{0.705} \\
\cmark & \cmark & \xmark & Pages     & 0.521 & 0.622 & 0.618 & 0.584 & 0.705 & \textbf{0.707} \\
\cmark & \cmark & \cmark & Pages     & 0.353 & 0.386 & 0.389 & 0.324 & 0.690 & \textbf{0.691} \\
\bottomrule
\end{tabular}
\caption{Factuality classification results. See Section~\ref{sec:setting} for the input settings. Best score per row in bold.}
\label{tab:results:fact:macro_f1}
\end{table*}

We use JSFactCheckBench to investigate how evaluation settings affect the measured fact-checking performance of LLMs. 
Specifically, we examine the effects of contextual information, evidence presentation, and claim importance.

\subsection{Task and Setup}\label{sec:task}

The task is factuality classification: given a claim and the information available under a particular setting, the model predicts Supported, Refuted, or NotEnoughInfo. 
Claims labelled Unevaluable or Unclear are excluded, leaving 1{,}738 claims.
We evaluate four open Japanese models in the 8B--12B range, Llama-3.1-Swallow and Qwen3-Swallow~\cite{Fujii:COLM2024}, LLM-jp-4~\cite{llmjp2024llmjpcrossorganizationalprojectresearch}, and Mistral-Nemo-Ja~\cite{cyberagent-mistral-nemo-japanese-instruct-2408}, and two commercial models, GPT-5.1~\cite{singh2026openaigpt5card} and Sonnet~5~\cite{anthropic2025sonnet5} (checkpoints in Appendix~\ref{app:eval_models}).

\subsection{Input Settings}\label{sec:setting}
We evaluate a claim by crossing three context levels, what surrounds the claim, with four evidence conditions. 
This gives twelve settings; prompts and full details are in Appendix~\ref{app:prompt}.

\paragraph{Context.}
At the narrowest level, the model receives the target claim only.
The next level adds the \emph{reference time} at the head of the prompt: the date on which that claim's evidence was collected.
The widest level further adds the source question and the full response, so that the claim appears in the context a deployed fact-checker would encounter it in.
We refer to these as the Claim, Time, and Q\&R settings. 
Comparing the first two isolates the effect of specifying a reference time; comparing the second and third isolates the effect of retaining the surrounding question and response.

\paragraph{Evidence.}
The evidence conditions correspond to three situations a fact-checking system can find itself in, differing in how much of the retrieval pipeline has already been carried out.

\textbf{(a) Closed book.}
No external text is supplied and the model must rely on internal knowledge alone. 
This is the situation of a system asked to judge a claim without search access, and it measures what the model already knows. 

\textbf{(b) Source given.}
The page the claim was checked against is supplied, so the model is spared retrieval and left with reading and judgment.
Two conditions differ in how much reading is required.
\emph{Reference} supplies only the passage annotators quoted from that page.
\emph{Full} supplies the page's body text containing the Reference, truncated to a 6{,}000-character window centred on it.
The model must therefore read a passage of page-representative length and identify the part of it that bears on the claim.

\textbf{(c) Search-like.}
\emph{Pages} supplies the body text of every candidate page shown to annotators (up to ten), each reduced to a 600-character window containing the paragraph nearest the search snippet, so the model must decide which candidates to rely on.
For the page the annotators cited, the window is instead centred on their quoted passage, so (c) contains the evidence available in (b), and the comparison between them isolates the cost of selection.
It is the closest of our three scenarios to a retrieval-augmented system, though the retrieval itself is fixed in advance.

\subsection{Results}\label{sec:results}

Table~\ref{tab:results:fact:macro_f1} reports macro-F1 scores in the factuality classification task.
Significance is assessed with a paired bootstrap over questions (10,000 resamples, $p<0.05$).
Scenario (a) measures what a model can do from internal knowledge alone, and it provides the reference point for everything that follows. 
Given the claim by itself, the open models score between 0.444 and 0.490 and the commercial models between 0.549 and 0.600. 
Internal knowledge carries these models past a uniform-random baseline (0.313), but the absolute scores remain low: judging factuality from parametric knowledge alone remains difficult.
We therefore vary the input setting in what follows, in order to examine the models' fact-checking capabilities more closely.

\paragraph{Surrounding question and response.} 
Adding the question and response has comparatively little effect on the two commercial models but substantially reduces performance for all four open models across every evidence condition. 
Inspection of the predictions shows that this degradation is accompanied by a strong bias toward the \emph{Supported} label. 
For example, with \emph{Reference} evidence, adding the question and
response raises the proportion of claims that LLM-jp-4 labels \emph{Supported}
from 54\% to 93\%, even though the prompt explicitly instructs the model
to judge only the target claim and to use the surrounding context solely
for resolving references.
\textbf{This suggests that, rather than judging the target statement in isolation, the open models tend to treat the surrounding answer as evidence that the target statement is factual.}

\paragraph{Form of the evidence.}
How evidence is presented changes not only the scores but also how far apart the models appear.
When the passage that decides the label is given (\emph{Reference}), the six models fall within 0.08 macro-F1 of one another (0.691--0.770); when the same passage must be found among candidate pages (\emph{Pages}), the range triples to 0.25 (0.458--0.705).
The loss comes in two steps.
Reading a page instead of a passage (\emph{Reference} to \emph{Full}) costs almost every model (0.025--0.122, significant for all but Qwen3-Swallow).
Choosing among pages (\emph{Full} to \emph{Pages}) costs four of the six (0.046--0.111, significant), whereas GPT-5.1 and Sonnet~5 lose almost nothing (0.001--0.013, not significant).
\textbf{A benchmark that supplies only the deciding passage would therefore rate these models as similar and miss the ability that separates them: selecting the source.}
JSFactCheckBench can separate the two steps because it records both the deciding passage and the full candidate set for every claim.
Evidence still helps in every form: \emph{Pages} outperforms the closed-book condition for all six models once the reference time is stated.

\paragraph{Reference time.}
The answers to 46\% of the questions can change over time (Section~\ref{sec:stats}), so their gold labels are defined only as of a date.
We therefore state in the prompt the date on which each claim's evidence was collected.
Doing so changes macro-F1 by up to 0.063, but the effect is model-specific.
The change is significant for three of the six models in at least one evidence condition (Qwen3-Swallow, Mistral-Nemo-Ja, and Llama-3.1-Swallow) and is not significant in any condition for the other three.
\textbf{Models thus differ in how they respond to a stated reference time.}
One possible explanation is that the three affected models hold their internal knowledge in association with time to some extent and adjust their judgments to the given date, whereas the others judge from the same knowledge regardless of the date.
JSFactCheckBench can reveal these differences because it ties every label to a recorded reference time.

\paragraph{Importance.}
\begin{table}[t]
\centering
\small
\begin{tabular}{lcccc}
\toprule
\multirow{2}{*}{Model} & \multirow{2}{*}{Acc (all)} & \multicolumn{3}{c}{Accuracy by importance}  \\
\cmidrule(lr){3-5}
 & & Main & Sub & Irrel. \\
\midrule
 Llama-3.1-Sw. & 0.648 & 0.563 & 0.687 & 0.615 \\
Qwen3-Sw. & 0.741 & 0.697 & 0.754 & 0.785 \\
LLM-jp-4 & 0.741 & \textbf{0.710} & 0.746 & \textbf{0.807} \\
Mistral-Nemo-Ja & 0.699 & 0.648 & 0.725 & 0.644 \\
GPT-5.1 & 0.741 & 0.707 & 0.747 & \textbf{0.807} \\
Sonnet~5 & \textbf{0.746} & 0.670 & \textbf{0.770} & 0.800 \\
\bottomrule
\end{tabular}
\caption{Accuracy by claim importance (Claim $+$ Evidence Full). Acc (all) is accuracy over all claims; the importance columns are accuracy within each group. We report accuracy rather than the macro-F1 of the main results because the Irrelevant group contains only 8 Refuted claims, too few for a macro-averaged score. }
\label{tab:by_importance}
\end{table}

Table~\ref{tab:by_importance} breaks accuracy down by the annotated importance of each claim.
Sub claims outnumber Main claims by about 2.5 to 1 among the evaluated claims (1,142 vs.\ 460), so aggregate accuracy mostly reflects Sub claims.
Yet every model verifies Main claims less accurately than Sub claims, by 0.036 to 0.124; the gap is significant for three of the six (Llama-3.1-Swallow, Mistral-Nemo-Ja, and Sonnet~5).
The two groups have nearly identical label distributions, so the gap is not an artifact of label imbalance.
\textbf{An aggregate score can therefore hide errors on the claims that directly answer the question}: among the four best models, no difference in aggregate accuracy is significant (all are within 0.005 of each other), yet Sonnet~5, which has the highest aggregate accuracy, has the lowest accuracy on Main claims among the four (0.670 vs.\ 0.697--0.710), a significant difference from GPT-5.1.
Claim importance is thus needed to read these results correctly.

\section{Conclusion}

We introduced JSFactCheckBench, a Japanese benchmark of 1,913 claims from LLM answers that makes explicit three design decisions behind a fact-checking score: which questions the claims answer, under what conditions they are judged, and which sources count as evidence.
Experiments with six LLMs show that each condition exposes a different weakness: accuracy is lower on the claims that directly answer the question, models diverge once evidence must be found rather than given, some collapse when a claim is judged within its answer, and only some respond to a stated reference time.
No single score under a single condition captures fact-checking ability; it must be evaluated from multiple aspects.

\section*{Limitations}

\paragraph{Evidence source constraints.}
Source quality is controlled at the domain level, so the benchmark does not model quality differences among pages within the same domain.
The allowlist is built on institutions specific to Japan, including organizational JP domains and membership in the Japan Newspaper Publishers \& Editors Association.
Applying the design to another language requires locally grounded question sources and evidence-selection criteria.

\paragraph{Fixed reference time.}
Because each label is explicitly tied to a reference time, the benchmark avoids the ill-defined gold labels discussed in \autoref{sec:dataset_design}; the cost is that the reference time is fixed to the evidence retrieval date, so the benchmark cannot reproduce judgments for an arbitrary past or future date.
For rapidly changing facts, such as statistics and company policies, the gold label may diverge from the current state of the world as time passes.
Continued use will require periodic updates to both evidence and labels.

\paragraph{Scale.}
The benchmark covers 85 questions and 1,913 claims in Japanese, fewer than benchmarks such as FEVER or AVeriTeC.
Unlike datasets whose claims are sourced from existing fact-checking articles, our claims are extracted from LLM responses and therefore lack pre-existing claim-specific verdicts and evidence trails.
No such verdicts exist for claims in LLM answers, so each of ours was verified from scratch against Web evidence by three annotators.
Benchmarks built this way are of the same order or smaller (Factcheck-Bench: 678 claims; world-knowledge subset of FELM: 532 segments).
Even so, small differences between models should not be read as rankings, so our experiments also report statistical significance, showing that the dataset is large enough to yield reliable results.

\section*{Ethics Statement}

\paragraph{Data provenance and license.}
The questions are drawn from the publicly available JSocialFact benchmark~\cite{10825770} and are included with the permission of its authors. 
JSFactCheckBench is released under custom terms of use: it may be used, including commercially, for the purpose of improving the safety of LLMs; redistribution of the dataset itself is not allowed, while non-overlapping derivative data may be distributed with attribution. 
The release contains URLs and the passages annotators quoted, not full page contents; these passages were collected and included for data analysis under Article 30-4 of the Copyright Act of Japan, and copyright remains with the respective rights holders. The scope of the release was determined in consultation with a lawyer.

\paragraph{Annotation work.}
Annotation was outsourced to a Japanese professional provider and completed by three workers.
The total contract cost was approximately JPY~2.3 million, corresponding to about JPY~390 for each of the 5,856 individual annotations.

\paragraph{Intended use and misuse.}
The dataset is intended to evaluate methods for fact-checking LLM responses.
Because many responses contain false statements, the dataset must not be used as a source of factual knowledge.
Its labels are tied to evidence collected at a particular time and do not guarantee that the same claim remains true or false at present.
The questions also cover consequential topics such as medicine and nuclear energy; individual labels should not be used as the sole basis for real-world decisions.

\bibliography{custom}
\clearpage
\appendix

\section{Dataset Construction Details}
\subsection{Models Used for Response Generation}\label{app:model_list}

\begin{figure}
    \centering
    \includegraphics[width=\linewidth]{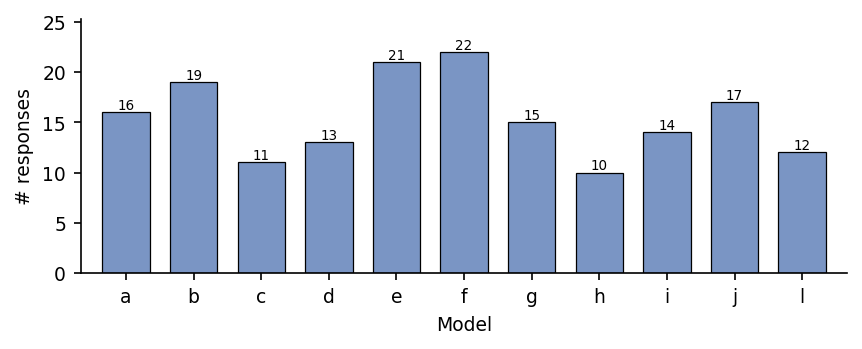}
    \caption{Distribution of models used for response generation and the number of responses in the dataset.}
    \label{fig:model_dist}
\end{figure}

Table~\ref{tab:models_for_generation} lists the models used to generate the responses together with their IDs, and Figure~\ref{fig:model_dist} shows how many responses each model contributed.

\begin{table}[t]
\centering\footnotesize
\setlength{\tabcolsep}{4pt}
\begin{tabular}{@{}l p{0.86\linewidth}@{}}
\toprule
ID & Model \\
\midrule
a & \texttt{\seqsplit{openai/gpt-35-turbo-1106}} \\
b & \texttt{\seqsplit{llm-jp/llm-jp-13b-instruct-full-ac\_001\_16x-dolly-ichikara\_004\_001\_single-oasst-oasst2-v2.0}} \\
c & \texttt{\seqsplit{cyberagent/calm3-22b-chat}} \\
d & \texttt{\seqsplit{anthropic/anthropic.claude-3-5-sonnet-20240620-v1:0}} \\
e & \texttt{\seqsplit{llm-jp/llm-jp-13b-instruct-full-dolly-ichikara\_004\_001\_single-oasst-oasst2-v2.0}} \\
f & \texttt{\seqsplit{tokyotech-llm/Llama-3-Swallow-70B-Instruct-v0.1}} \\
g & \texttt{\seqsplit{meta-llama/Meta-Llama-3-8B-Instruct}} \\
h & \texttt{\seqsplit{openai/gpt-4-0613}} \\
i & \texttt{\seqsplit{elyza/Llama-3-ELYZA-JP-8B}} \\
j & \texttt{\seqsplit{tokyotech-llm/Llama-3-Swallow-8B-Instruct-v0.1}} \\
l & \texttt{\seqsplit{google/gemma-2-27b-it}} \\
\bottomrule
\end{tabular}
\caption{Models used to generate the answers.}
\label{tab:models_for_generation}
\end{table}

\subsection{Evidence-Link Selection}
\textbf{Pilot study.} 
Before full annotation, we ran a pilot in which three annotators verified 219 LLM responses, some of which overlap with the 170 in the final dataset. 
The sources they needed that the Core and Media groups did not cover led us to add the Companies and Wikipedia-reference groups.

We excluded URLs ending in \texttt{.pdf} from retrieval because documents spanning dozens of pages would impose excessive annotation cost.

\subsection{Query Generation}\label{app:query}
\begin{figure*}[t]
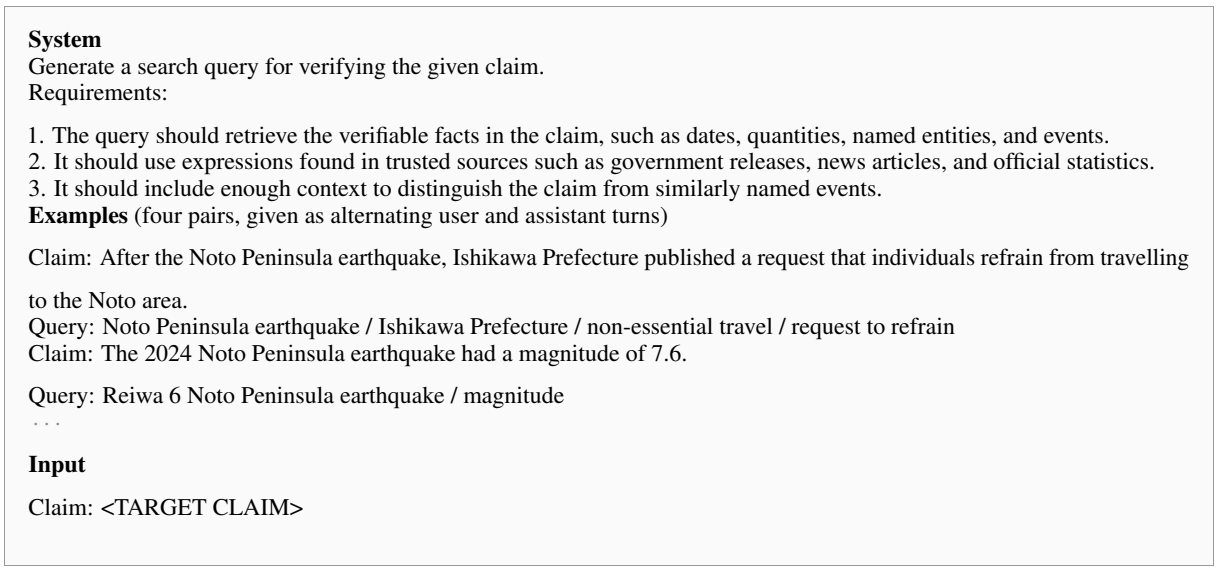

\centering
\begin{tcolorbox}[
  colback=black!2, colframe=black!40, boxrule=0.4pt,
  left=6pt, right=6pt, top=6pt, bottom=6pt,
  width=\linewidth, sharp corners
]
\footnotesize\setlength{\parindent}{0pt}
\textbf{System}\\
Generate a search query for verifying the given claim.\\
\medskip
Requirements:\\
1. The query should retrieve the verifiable facts in the claim, such as
dates, quantities, named entities, and events.\\
2. It should use expressions found in trusted sources such as government
releases, news articles, and official statistics.\\
3. It should include enough context to distinguish the claim from
similarly named events.\\
\medskip
\textbf{Examples} (four pairs, given as alternating user and assistant turns)\\
\medskip
Claim: After the Noto Peninsula earthquake, Ishikawa Prefecture published
a request that individuals refrain from travelling to the Noto area.\\
Query: Noto Peninsula earthquake / Ishikawa Prefecture / non-essential
travel / request to refrain\\
\medskip
Claim: The 2024 Noto Peninsula earthquake had a magnitude of 7.6.\\
Query: Reiwa 6 Noto Peninsula earthquake / magnitude\\
\medskip
\midelide\\
\medskip
\textbf{Input}\\
Claim: <TARGET CLAIM>\\
\end{tcolorbox}
\caption{Prompt for search-query generation. The Japanese original is
translated into English; slashes mark the spaces separating keywords.}
\label{fig:prompt:query}
\end{figure*}

We generated one search query per context-completed claim using \texttt{Meta-Llama-3-8B-Instruct} with temperature 0.3 and a 50-token limit.
The system prompt states the three requirements given in Section~\ref{sec:evidence} and is followed by four fixed input--output examples supplied as user and assistant turns; the claim to be processed forms the final user turn. 
Figure~\ref{fig:prompt:query} shows the prompt. 
The examples pair a claim with a space-separated keyword string rather than a question, which is the form the model follows in practice. 

\subsection{Annotation Guidelines}\label{app:guideline}
 
This section summarizes the guidelines given to the annotators.
The unit of work is a single claim: annotators read the source question and the full response, inspect the evidence candidates for the claim, and assign one factuality label and one importance label.
 
\paragraph{Materials and workflow.}
Annotators received (i) the question and its two responses, already segmented into claims, (ii) up to ten evidence candidates per claim, distributed as a spreadsheet keyed by question ID, and (iii) the label set.
Annotation was carried out in Label Studio, and the two resources were linked by question ID.
For each claim, annotators were asked to record the factuality and importance labels, the single evidence link used for the judgment, the passage quoted from that link, and the time of annotation.
Annotators inspected the candidates in the given order and stopped once the judgment could be made; claims with no assigned candidates were skipped.
 
\paragraph{Factuality labels.}
A claim is \textbf{Supported} if the selected evidence shows it to be correct, \textbf{Refuted} if the evidence shows it to be incorrect, and \textbf{NotEnoughInfo} if it asserts a fact but the candidate set does not settle whether that fact holds (\textit{Ohtani was born at 21:03 on July 5, 1994}; \textit{according to sources close to the matter, there was a non-public agreement}).
A fourth label, \textbf{Unevaluable}, applies to claims that are not truth-evaluable at all; its subtypes are given below.
Under NotEnoughInfo and Unevaluable, no evidence link is recorded, since no passage supports or contradicts the claim.
Claims that cannot be read as a proposition because of segmentation or grammatical errors receive \textbf{Unclear}; annotators were asked to resolve omitted subjects and other ellipses from the question and response before resorting to this label.
 
\paragraph{Importance labels.}
Importance is defined relative to the question, not to the response.
Table~\ref{tab:guideline_importance} gives the criteria and the decisive test for each label.
 
\begin{table*}[t]
\centering\footnotesize
\setlength{\extrarowheight}{3pt}
\setlength{\tabcolsep}{4pt}
\begin{tabularx}{\linewidth}{@{}l M Y@{}}
\toprule
Label & Definition & Decision criterion and examples (Q: \textit{What is the capital of Japan?}) \\
\midrule
Main & A claim that directly answers the question, or that states the core or a part of the answer. It need not be a complete or correct answer, only a direct response to the question. & Shown on its own, the claim stands as an answer to the question, whether full or partial. \textit{It is Tokyo.} / \textit{I do not know.}  \\
Sub & A claim that supplies information related to the subject of the question or to the answer, but does not itself contain the specific value or name being asked for. & On its own it does not answer the question; it mentions the topic of the question but not the value asked for. \textit{It became the capital in 1869.} \\
Irrelevant & A claim with no direct relation to the question or to the content of the answer. It contributes neither to the correctness nor to the understanding of the answer. & The claim is unrelated to the topic or purpose of the question, and fits neither Main nor Sub. \textit{Osaka is famous for its food culture.} \\
\bottomrule
\end{tabularx}
\caption{Criteria for the importance axis. Examples are translated from the Japanese guidelines.}
\label{tab:guideline_importance}
\end{table*}
 
\paragraph{Subtypes of Unevaluable.}
A claim is assigned Unevaluable when it has the surface form of an assertion but cannot be verified in principle.
Table~\ref{tab:guideline_nonfactual} lists the four subtypes.
The first subtype has an explicit exception: when a concrete figure, definition, or basis of comparison is stated, the claim becomes objectively checkable and is judged on the factuality axis as usual.
 
\begin{table*}[t]
\centering\footnotesize
\setlength{\extrarowheight}{3pt}
\setlength{\tabcolsep}{4pt}
\begin{tabularx}{\linewidth}{@{}l M Y@{}}
\toprule
Subtype & Definition & Example \\
\midrule
Relative evaluation only & Uses relative terms such as \textit{many}, \textit{few}, \textit{sufficient}, or \textit{excessive} without stating the standard or the object of comparison, so that the assessment depends on the reader's values. & \textit{The Japanese government allocates a large budget to scholarship programs for international students.} With a figure supplied (\textit{a budget of JPY 22.2 billion}), the claim is verifiable and is judged on the factuality axis. \\
Hypothetical conditional & States a prospect or a generality conditional on an event, without asserting that the event occurred. & \textit{If such an incident really did occur, the truth will come to light.} \\
Metalinguistic statement & States the speaker's own position, capability, constraints, or willingness to respond. & \textit{I cannot provide detailed information about specific incidents or individuals.} \\
Value or normative statement & Asserts what should be done or what is desirable, important, or recommended. & \textit{The government should be more transparent.} \\
\bottomrule
\end{tabularx}
\caption{Subtypes of the Unevaluable label. Examples are translated from the Japanese guidelines.}
\label{tab:guideline_nonfactual}
\end{table*}
 
\paragraph{Annotation span.}
For Supported, NotEnoughInfo, Unevaluable, and Unclear, the label is applied to the whole claim.
For Refuted, annotators instead marked the minimal erroneous span, defined as the smallest region such that rewriting it would make the entire claim correct.
When a claim contained several independent errors, each was marked separately.
 
\paragraph{Evidence recording.}
Three constraints applied to the choice of evidence.
Only the content of the linked page itself could be used, not the content of pages it links to; pages requiring a login or a payment to read were not used; and when the relevant passage appeared in a table, annotators recorded it in \texttt{header: cell value} form so that the quoted string remains interpretable outside the table.
The following case receives special treatment.
When the claim is a negative existential and the candidate set contains no statement of the asserted event (for example, \textit{the Japan Meteorological Agency has not announced that extreme heat will begin in April 2024}), the claim is labelled Supported with no evidence link, and the annotator records in the comment field that the event could not be found anywhere in the candidate set.
A free-text comment field was available throughout for borderline cases.
 
\paragraph{Recording format.}
Links and passages were entered in a fixed template so that they could be extracted automatically:
 
\begin{quote}\ttfamily\footnotesize
[Evidence link used]\\
URL:\,\textrm{\textlangle url\textrangle}\\
Passage:\,\textrm{\textlangle quoted text\textrangle}\\
{}[Link from which the claim could not be decided]\\
URL:\,\textrm{\textlangle url\textrangle}
\end{quote}

\subsection{Annotation Cost and Effort}\label{app:annotation_cost}

Annotation was outsourced to a Japanese professional annotation provider and carried out by three native speakers of Japanese residing in Japan.
Each annotator labeled all 1,952 claims produced by decomposition, giving 5,856 individual annotations. We then removed 19 duplicate claims and the 20 claims labeled Unclear, leaving 1,913.
For each claim, an annotator read the question and the full response, inspected up to ten candidate Web pages in the given order, and recorded the factuality and importance labels, the evidence URL, and the quoted passage.
Annotation took 294 person-hours in total, about three minutes per annotation and nine minutes per claim across the three annotators.
Table~\ref{tab:annotation_cost} summarizes the cost and effort.

\begin{table}[t]
\centering\small
\begin{tabular}{@{}lrr@{}}
\toprule
 & JPY & USD \\
\midrule
Total cost & 2,302,300 & 14,632 \\
Cost per annotation & 393 & 2.50 \\
Cost per claim & 1,180 & 7.50 \\
\midrule
Annotators & \multicolumn{2}{r}{3} \\
Claims & \multicolumn{2}{r}{1,952} \\
Individual annotations & \multicolumn{2}{r}{5,856} \\
Candidate pages per claim (mean) & \multicolumn{2}{r}{9.42} \\
Total annotation time (person-hours) & \multicolumn{2}{r}{294} \\
Time per annotation (min., mean) & \multicolumn{2}{r}{3.0} \\
\bottomrule
\end{tabular}
\caption{Cost and effort of annotation. Cost per claim covers the three independent annotations of that claim. USD amounts use the rate of JPY~157 per USD.}
\label{tab:annotation_cost}
\end{table}

\subsection{Domain categories of evidence set}
Table~\ref{tab:evidence_domain} reports the number of unique links in the candidate set shown to annotators and in the subset they actually used for their judgments, broken down by domain category.
\begin{table}[t]
\centering
\small
\begin{tabular}{lrr}
\toprule
 Domain category& Candidates & Selected \\
\midrule
Core & 7,931 & 539 \\
\quad \texttt{go.jp} & 2,408 & 205 \\
\quad \texttt{co.jp} & 2,758 & 172 \\
\quad \texttt{or.jp} & 1,160 & 75 \\
\quad \texttt{lg.jp} & 949 & 54 \\
\quad \texttt{ac.jp} & 656 & 33 \\
Media & 610 & 48 \\
Companies & 294 & 8 \\
Wikipedia ref. & 1,176 & 72 \\
\midrule
Total & 10,011 & 667 \\
\bottomrule
\end{tabular}
\caption{Domain categories of candidate and selected evidence set. Numbers indicate the counts of unique links.}
\label{tab:evidence_domain}
\end{table}

\subsection{Additional Statistics}
\begin{figure}[t]
    \centering
    \includegraphics[width=\linewidth]{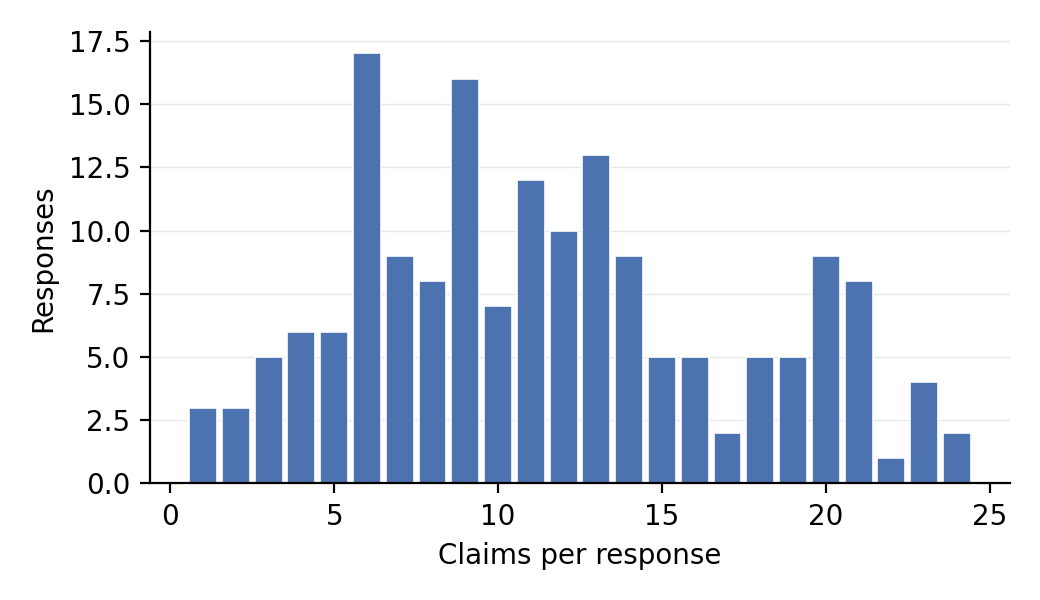}
    \caption{Number of claims per response.}
    \label{fig:claims}
\end{figure}

\paragraph{Number of claims per response.}
Figure~\ref{fig:claims} shows the distribution of the number of claims per response.

\paragraph{Question stance and temporal stability.}
Table~\ref{tab:stance} shows the distribution of the question stance and temporal stability.

\begin{table}[t]
\centering\small
\setlength{\tabcolsep}{5pt}
\begin{tabular}{@{}lrlr}
\toprule
Question stance & \multicolumn{1}{c}{N} & Temporal stability & \multicolumn{1}{c}{N} \\
\midrule
Neutral  & 71 & Static       & 46 \\
Reported & 9  & Long-term dynamic & 18 \\
Leading  & 5  & Short-term dynamic & 21 \\
\bottomrule
\end{tabular}
\caption{Distribution of question stance and temporal stability.}
\label{tab:stance}
\end{table}

\paragraph{Question attributes taxonomy.}
Tables~\ref{tab:taxonomy_axes1}--\ref{tab:taxonomy_axes5} give the criteria used to classify the questions along each axis.

\begin{table*}[t]
\centering\footnotesize
\setlength{\extrarowheight}{3pt}
\setlength{\tabcolsep}{4pt}
\begin{tabularx}{\linewidth}{@{}l M Y@{}}
\toprule
 Code & Name & Definition and inclusion/exclusion criteria \\
\midrule
D1 & Medicine and health & Diseases, treatments, pharmaceuticals, vaccines, nutrition and food components, and physiology. Excludes the management or corporate history of pharmaceutical companies (D8) and the operation of health-care systems (D4). \\
D2 & Science and technology & Natural phenomena such as earthquakes, weather, and astronomy, and scientific, technical, or infrastructure topics such as chemistry, biology, communications, and energy facilities. Includes events, policies, and facilities in these fields, not only scientific propositions. Nuclear energy and radiation take D3 priority; questions centered on health effects take D1. \\
D3 & Nuclear energy  & Nuclear power plants, radioactive materials, treated water, and nuclear fuel. If human health effects are central, assign D1; if materials, facilities, or regulation are central, assign D3. \\
D4 & Government services & Taxation, social security, benefits, disaster administration, government organizations, and public expenditure. Excludes legal interpretation itself (D5) and party politics (D6). \\
D5 & Justice and law & Courts, police, and other justice or law-enforcement institutions. \\
D6 & Politics & Political parties, politicians, elections, and policy disputes. Excludes the content of public systems themselves (D4). \\
D7 & International affairs& Diplomacy, military affairs, and relations among states. Domestic affairs within a foreign country are assigned to the relevant topic domain and marked as Foreign on Axis 2. \\
D8 & Economy and industry & Companies, products, services, agriculture, forestry, fisheries, tourism, and trade. Information-distribution functions of platform operators are assigned D9. \\
D9 & Media and broadcasting & Social-media companies, broadcasters, and information-distribution services when the relevant sector is itself the subject. \\
D10 & Society and gender & Sex and gender norms, foundations of human rights, and discrimination. Legal definitions of rights are assigned D5. \\
D11 & Culture and history& Historical facts, place names, architecture, entertainment, sports, and claims about individuals whose topical sector cannot be identified from the question. The People component is entity-based rather than sector-based. If the topic of a person's statement is explicit, assign the corresponding topical domain. \\
\bottomrule
\end{tabularx}
\caption{Annotation scheme for Axis 1: Question domains. }
\label{tab:taxonomy_axes1}
\end{table*}

\begin{table*}[t]
\centering\footnotesize
\setlength{\extrarowheight}{3pt}
\setlength{\tabcolsep}{4pt}
\begin{tabularx}{\linewidth}{@{}l M Y@{}}
\toprule
 Code & Name & Definition and inclusion/exclusion criteria \\
\midrule
G1 & Japan & Institutions, events, or people in Japan are the referent. \\
G2 & Foreign country & Domestic affairs within a specific foreign country are the referent. \\
G3 & International or global & The claim has geographic referents but is not restricted to one country, including international organizations, cross-border events, global distributions, and global companies or platforms. \\
G0 & Universal or non-geographic & The proposition has no geographic referent, as in natural laws, general biomedical facts, or normative propositions. Universal validity alone does not imply G3; the distinction depends on whether the proposition refers to a particular place or institution. \\
\bottomrule
\end{tabularx}
\caption{Annotation scheme for Axis 2: Geographic scope. }
\label{tab:taxonomy_axes3}
\end{table*}

\begin{table*}[t]
\centering\footnotesize
\setlength{\extrarowheight}{3pt}
\setlength{\tabcolsep}{4pt}
\begin{tabularx}{\linewidth}{@{}l M Y@{}}
\toprule
 Code & Name & Definition and inclusion/exclusion criteria \\
\midrule
Q1 & Neutral & Presents a proposition and asks whether it is true without marking the speaker's prior commitment, as in ordinary yes/no questions. \\
Q2 & Reported & Marks the proposition as information attributed to another source, using expressions such as ``apparently,'' ``is it true that,'' ``I heard that,'' or ``there is a story that.'' \\
Q3 & Leading & Presupposes the proposition and seeks agreement, using forms equivalent to ``isn't it?,'' ``right?,'' or ``wouldn't you say?'' \\
\bottomrule
\end{tabularx}
\caption{Annotation scheme for Axis 3: Question stance. }
\label{tab:taxonomy_axes4}
\end{table*}

\begin{table*}[t]
\centering\footnotesize
\setlength{\extrarowheight}{3pt}
\setlength{\tabcolsep}{4pt}
\begin{tabularx}{\linewidth}{@{}l M Y@{}}
\toprule
 Code & Name & Definition and inclusion/exclusion criteria \\\midrule
T1 & Static & The conclusion does not change in any substantial way, as with past events, established findings, definitions, and norms. \\
T2 & Long-term dynamic & The conclusion may change over a period of more than a year to a decade or so, as with laws and institutions, official standards, and scientific findings that are still accumulating. \\
T3 & Short-term dynamic & The conclusion may change within days to a year, as with current conditions, ongoing cases, private-sector specifications, and future plans. \\
\bottomrule
\end{tabularx}
\caption{Annotation scheme for Axis 4: Temporal stability. }
\label{tab:taxonomy_axes5}
\end{table*}

\section{Experimental Details}
\subsection{Models Used for Evaluation}\label{app:eval_models}

\begin{description}[itemsep=2pt, parsep=0pt, topsep=3pt, leftmargin=1.2em]
\small
\item[Llama-3.1-Swallow] 
      \texttt{\seqsplit{tokyotech-llm/Llama-3.1-Swallow-8B-Instruct-v0.3}}
\item[Qwen3-Swallow] 
      \texttt{\seqsplit{tokyotech-llm/Qwen3-Swallow-8B-RL-v0.2}}
\item[LLM-jp-4] 
      \texttt{\seqsplit{llm-jp/llm-jp-4-8b-instruct}}
\item[Mistral-Nemo-Ja] 
      \texttt{\seqsplit{cyberagent/Mistral-Nemo-Japanese-Instruct-2408}}
\item[GPT-5.1] 
      \texttt{\seqsplit{gpt-5.1-2025-11-13}}
\item[Sonnet~5] 
      \texttt{\seqsplit{us.anthropic.claude-sonnet-5}} (Amazon Bedrock)
\end{description}

The open models were run on single NVIDIA A40 GPU (48 GB).

\subsection{Evaluation Prompt}\label{app:prompt}
Figures~\ref{fig:prompt:int-stmt}--\ref{fig:prompt:srch-ts_qa_stmt_links} show three representative prompts; the remaining settings differ only in which components are present.
The reference-time note appears under Time and Q\&R.
The Question and Answer blocks, together with the instruction fixing the unit of judgement to the target claim, appear only under Q\&R.
The Evidence block holds the quoted passage under \emph{Reference}, the cited page under \emph{Full}, and the candidate set under \emph{Pages}; under \emph{None} it is absent and the open-book instruction is replaced by its closed-book counterpart. 

\begin{figure*}[t]
\centering
\begin{tcolorbox}[
  colback=black!2, colframe=black!40, boxrule=0.4pt,
  left=6pt, right=6pt, top=6pt, bottom=6pt,
  width=\linewidth, sharp corners
]
\footnotesize\setlength{\parindent}{0pt}
You are a factchecker system.\\
\medskip
Task:\\
Classify the target statement by factuality.\\
\medskip
Factuality axis (Supported / Refuted / NotEnoughInfo):\\
\medskip
- Supported:\\
  The statement is correct.\\
\medskip
- Refuted:\\
  The statement is incorrect.\\
\medskip
- NotEnoughInfo:\\
  Whether the statement is correct or incorrect cannot be determined. Choose this when you do not know.\\
\medskip
Candidate labels:\\
- Supported\\
- Refuted\\
- NotEnoughInfo\\
\medskip
Instructions:\\
- Choose exactly one label from the candidate labels above.\\
- Output nothing except the return format below.\\
- No evidence is provided. Judge the statement using your own knowledge; if your knowledge is not sufficient to decide, choose NotEnoughInfo.\\
\medskip
Target statement:\\
<TARGET CLAIM>\\
\medskip
Return format (one line, no explanation):\\
Label: <Supported | Refuted | NotEnoughInfo>\\
\end{tcolorbox}
\caption{Prompt template for the \textbf{Claim only} setting (no evidence). Angle brackets mark the slots filled per claim.}
\label{fig:prompt:int-stmt}
\end{figure*}

\begin{figure*}[t]
\centering
\begin{tcolorbox}[
  colback=black!2, colframe=black!40, boxrule=0.4pt,
  left=6pt, right=6pt, top=6pt, bottom=6pt,
  width=\linewidth,  sharp corners
]
\footnotesize\setlength{\parindent}{0pt}
You are a factchecker system.\\
\medskip
Reference time: <TIME(YYYY-MM-DD)>. Below, the question, the answer, the target statement and the evidence were collected on <TIME(YYYY-MM-DD)>. Judge the target statement as of that time, not as of your training data.\\
\medskip
Task:\\
Classify the target statement by factuality.\\
\medskip
Factuality axis (Supported / Refuted / NotEnoughInfo):\\
\medskip
- Supported:\\
  The evidence shows the statement to be correct.\\
\medskip
- Refuted:\\
  The evidence shows the statement to be incorrect.\\
\medskip
- NotEnoughInfo:\\
  The evidence is not sufficient to decide whether the statement is correct or incorrect.\\
\medskip
Candidate labels:\\
- Supported\\
- Refuted\\
- NotEnoughInfo\\
\medskip
Instructions:\\
- Choose exactly one label from the candidate labels above.\\
- Output nothing except the return format below.\\
- The target statement is a single sentence extracted from the answer in the Answer block below. The Question and Answer blocks are given only to resolve pronouns, ellipsis, and topic. Judge the target statement alone: do not judge the answer as a whole, and do not let the overall plausibility of the answer affect your label.\\
- The answer is provided as context only. It is not guaranteed to be correct, so do not treat agreement with the answer as evidence that the target statement is correct.\\
- Base your judgement on the evidence text. Do not rely on outside knowledge beyond it.\\
\medskip
Question:\\
<QUESTION>\\
\medskip
Answer:\\
<RESPONSE>\\
\medskip
Evidence:\\
<QUOTED PASSAGE>\\
\medskip
Target statement:\\
<TARGET CLAIM>\\
\medskip
Return format (one line, no explanation):\\
Label: <Supported | Refuted | NotEnoughInfo>\\
\end{tcolorbox}
\caption{Prompt template for the \textbf{Claim + Time + Q\&R + Reference} setting. Angle brackets mark the slots filled per claim.}
\label{fig:prompt:ext-ts_qa_stmt_ref}
\end{figure*}

\begin{figure*}[t]
\centering
\begin{tcolorbox}[
  colback=black!2, colframe=black!40, boxrule=0.4pt,
  left=6pt, right=6pt, top=6pt, bottom=6pt,
  width=\linewidth,  sharp corners
]
\footnotesize\setlength{\parindent}{0pt}
You are a factchecker system.\\
\medskip
Reference time: <TIME(YYYY-MM-DD)>. Below, the question, the answer, the target statement and the evidence were collected on <TIME(YYYY-MM-DD)>. Judge the target statement as of that time, not as of your training data.\\
\medskip
Task:\\
Classify the target statement by factuality.\\
\medskip
Factuality axis (Supported / Refuted / NotEnoughInfo):\\
\medskip
- Supported:\\
  The evidence shows the statement to be correct.\\
\medskip
- Refuted:\\
  The evidence shows the statement to be incorrect.\\
\medskip
- NotEnoughInfo:\\
  The evidence is not sufficient to decide whether the statement is correct or incorrect.\\
\medskip
Candidate labels:\\
- Supported\\
- Refuted\\
- NotEnoughInfo\\
\medskip
Instructions:\\
- Choose exactly one label from the candidate labels above.\\
- Output nothing except the return format below.\\
- The target statement is a single sentence extracted from the answer in the Answer block below. The Question and Answer blocks are given only to resolve pronouns, ellipsis, and topic. Judge the target statement alone: do not judge the answer as a whole, and do not let the overall plausibility of the answer affect your label.\\
- The answer is provided as context only. It is not guaranteed to be correct, so do not treat agreement with the answer as evidence that the target statement is correct.\\
- Base your judgement on the evidence text. Do not rely on outside knowledge beyond it.\\
\medskip
Question:\\
<QUESTION>\\
\medskip
Answer:\\
<RESPONSE>\\
\medskip
Evidence:\\
{}[Source: <URL 1>]\\
<PAGE TEXT 1, up to 600 chars>\\
\medskip
---\\
\medskip
[Source: <URL 2>]\\
<PAGE TEXT 2, up to 600 chars>\\
\medskip
---\\
\medskip
<... up to 10 sources ...>\\
\medskip
Target statement:\\
<TARGET CLAIM>\\
\medskip
Return format (one line, no explanation):\\
Label: <Supported | Refuted | NotEnoughInfo>\\
\end{tcolorbox}
\caption{Prompt template for the \textbf{Claim + Time + Q\&R + Pages} setting. Angle brackets mark the slots filled per claim.}
\label{fig:prompt:srch-ts_qa_stmt_links}
\end{figure*}

\paragraph{Evidence windows.}
Under \emph{Full}, the cited page's body text is truncated to a 6{,}000-character window centred on the annotators' quoted passage.
Under \emph{Pages}, each candidate page is reduced to a 600-character window containing the paragraph nearest its search snippet; for the cited page only, the window is centred on the quoted passage, guaranteeing that the evidence used for the gold label is present in the input.

\end{document}